\documentclass[conference]{IEEEtran}
\IEEEoverridecommandlockouts
\usepackage{cite}
\usepackage{hyperref}
\usepackage{amsmath,amssymb,amsfonts}
\usepackage{algorithmic}
\usepackage{booktabs}
\usepackage{makecell}
\usepackage{multirow}
\usepackage{pifont}
\usepackage[normalem]{ulem}
\usepackage{mathabx}
\usepackage{arydshln}
\usepackage{multirow}
\usepackage{graphicx}
\usepackage{textcomp}
\usepackage{xcolor}
\usepackage{subcaption}
\def\BibTeX{{\rm B\kern-.05em{\sc i\kern-.025em b}\kern-.08em
    T\kern-.1667em\lower.7ex\hbox{E}\kern-.125emX}}
\begin{document}

\title{Serial Low-rank Adaptation of Vision Transformer}

\author{
    \IEEEauthorblockN{Houqiang Zhong$^{1}$\IEEEauthorrefmark{1}, Shaocheng Shen$^{2}$\IEEEauthorrefmark{1}, Ke Cai$^{3}$\IEEEauthorrefmark{1}, Zhenlong Wu$^{1}$, Jiangchao Yao$^{2}$, Yuan Cheng$^{3}$}
    \IEEEauthorblockN{Xuefei Li$^{3}$, Xiaoyun Zhang$^{2}$, Li Song$^{1}$\IEEEauthorrefmark{2}, Qiang Hu$^{2}$\IEEEauthorrefmark{2}}
    \IEEEauthorblockA{$^1$ School of Information Science and Electronic Engineering, Shanghai Jiao Tong University, Shanghai, China}
    \IEEEauthorblockA{$^2$ Cooperative Medianet Innovation Center, Shanghai Jiao Tong University, Shanghai, China}
    \IEEEauthorblockA{$^3$ Glodon Company, Shanghai, China}
    \thanks{\IEEEauthorrefmark{1} Equal contributions. \IEEEauthorrefmark{2} Li Song(song\_li@sjtu.edu.cn) and Qiang Hu (qiang.hu@sjtu.edu.cn) are the corresponding authors. This work is supported by National Natural Science Foundation of China (62271308), STCSM (24ZR1432000, 24511106902, 24511106900, 22511105700, 22DZ2229005), 111 plan  (BP0719010).}
}

\maketitle

\begin{abstract}

Fine-tuning large pre-trained vision foundation models in a parameter-efficient manner is critical for downstream vision tasks, considering the practical constraints of computational and storage costs. Low-rank adaptation (LoRA) is a well-established technique in this domain, achieving impressive efficiency by reducing the parameter space to a low-rank form. However, developing more advanced low-rank adaptation methods to reduce parameters and memory requirements remains a significant challenge in resource-constrained application scenarios. In this study, we consider on top of the commonly used vision transformer and propose Serial LoRA, a novel LoRA variant that introduces a shared low-rank matrix serially composite with the attention mechanism. Such a design extracts the underlying commonality of parameters in adaptation, significantly reducing redundancy. Notably,  Serial LoRA uses only \textcolor{magenta}{1/4} parameters of LoRA but achieves comparable performance in most cases. We conduct extensive experiments on a range of vision foundation models with the transformer structure, and the results confirm consistent superiority of our method.

\end{abstract}    
\section{Introduction}
\label{sec:intro}

Vision foundation models, such as SAM~\cite{Kirillov_2023_ICCV}, CLIP~\cite{pmlr-v139-radford21a}, and diffusion models~\cite{SD3}, particularly those based on large-scale transformer architectures, have driven rapid advancements in computer vision, achieving breakthroughs in tasks like classification, segmentation and image generation. Despite the impressive performance, deploying these large-scale pre-trained models in real-world applications faces inevitable limitations, especially in resource-constrained environments like mobile devices and edge computing platforms, where high computational and memory demands present challenges. Consequently, parameter-efficient fine-tuning (PEFT) methods are developed to reduce storage and computational costs, making vision foundation models more accessible.

Current PEFT methods encompass different paradigms like prompt tuning \cite{zhu2024minigpt,dai2023instructblip,Benchmarking}, adapter-based methods \cite{huetal2023llm,adapterfusion,AdapterSoup} and low-rank adaptation (LoRA) \cite{hu2022lora,li2024loftq,hayou2024loraefficientlowrank}, where we place focus on the latter in this study, given its potential in efficiency and parameter size reduction. As a representative approach, LoRA constructs the product of two low-rank matrices to narrow down the optimization space, making it possible to adapt vision foundation models to new tasks with fewer resources. 
For example, recent studies~\cite{Peng_2024_CVPR,li2024adapting} have shown that LoRA can effectively adapt vision transformer blocks in SAM for precise anatomical segmentation, while significantly reducing the parameter overhead in domain adaptation.
In text-to-image generation, recent works~\cite{LCMLORA,Hartley_2024_CVPR,smith2024continual} have successfully integrated LoRA into diffusion models, enabling efficient personalization of generative models while maintaining high-fidelity image synthesis capabilities. Despite effectiveness, current LoRA has not well considered the structural characteristic of vision transformer.
This makes us rethink how to better align PEFT methods with the architectural nuances of vision transformers, ensuring scalability and efficiency in fine-tuning large-scale vision models.

\begin{figure}[t!]
    \centering
    \setlength{\belowcaptionskip}{-0.3cm}
    \includegraphics[width=\linewidth]{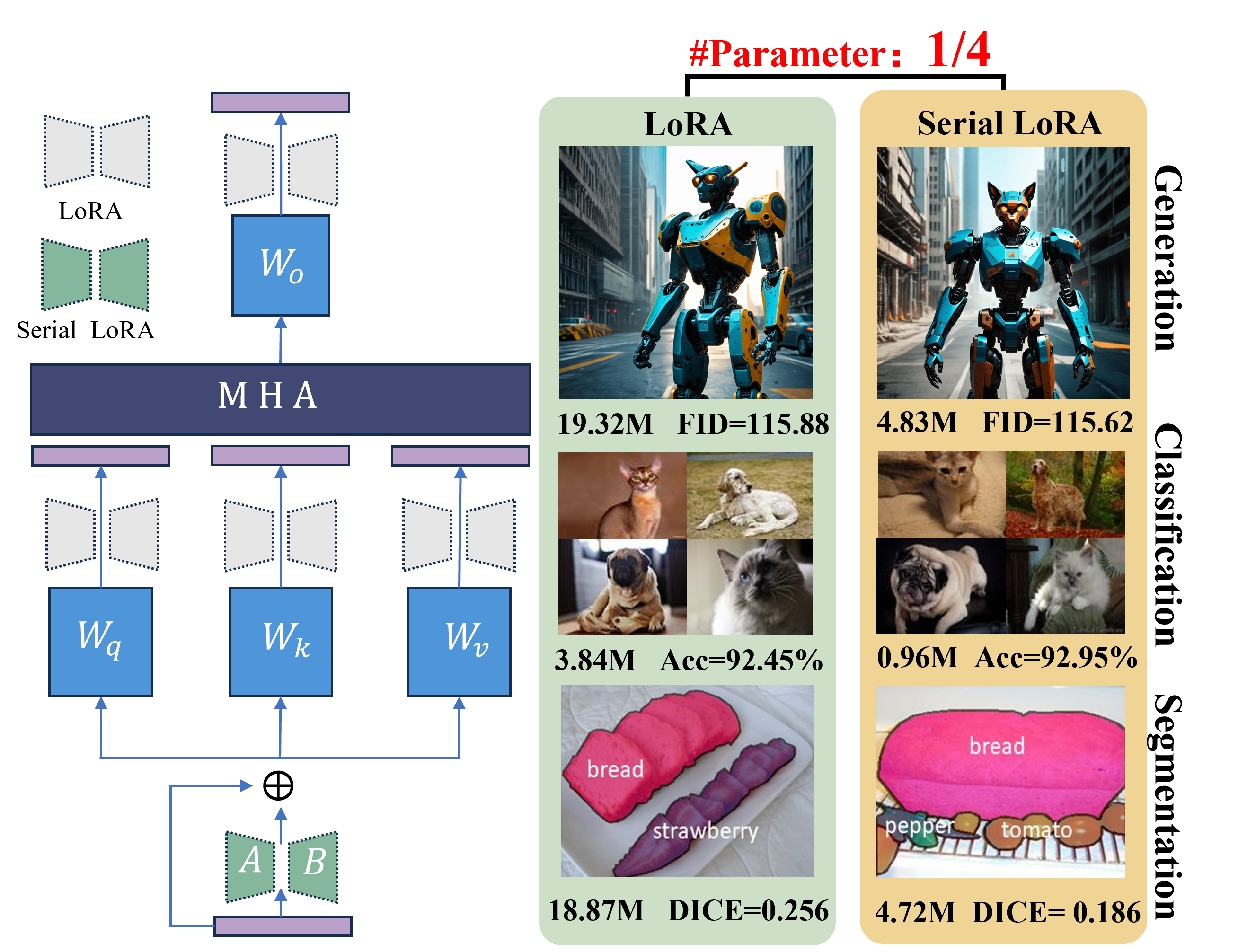}
    \caption{\textbf{Left}: Serial LoRA structure, illustrating the application of Serial LoRA in Transformers. \textbf{Right}: Comparison results of Serial LoRA and LoRA across various computer vision tasks, including image generation, image classification and semantic segmentation. \vspace{1em}}
    \label{fig:teaser}
\end{figure}
Generally, in vision transformers, the multi-head attention (MHA) mechanism and the QKV design tend to use larger feature dimensions and more attention heads to enhance the model's ability to capture relationships across multiple feature spaces. However, the fine-tuning process must adjust parameters across the individual feature spaces of Q, K, V, and O for each attention head. this parallel structure leads to an increase in fine-tuning parameters, making adaptation more computationally expensive. This paper explores a fundamental question: Is there a shared low-rank LoRA space that can efficiently adapt the transformer structure? We hypothesize that such a unified LoRA space could capture underlying parameter commonalities across feature subspaces, reducing redundancy and the number of parameters required for fine-tuning. This concept serves as the foundation that motivates us to propose a more efficient LoRA method to finetuning vision foundation models with the transformer structure.

To validate this hypothesis, we propose Serial LoRA, a novel variant specifically designed to address the parameter-intensive parallel structure in Transformers. Serial LoRA employs a shared low-rank matrix that can be serially composite with the attention mechanism, as shown in Fig. \ref{fig:teaser}, enabling the attention heads to efficiently share a unified low-rank space. This design extracts underlying parameter commonalities across heads, significantly reducing redundancy and fine-tuning costs. Our comprehensive evaluations across various transformer-based vision foundation models demonstrate the effectiveness of Serial LoRA. Specifically, we conduct extensive experiments on 24 datasets spanning three major vision tasks: image classification using CLIP, semantic segmentation with SAM, and image generation based on Stable Diffusion 3. The results consistently show that Serial LoRA achieves comparable performance while reducing model parameters by 75\% compared to standard LoRA. Furthermore, we successfully integrate learning rate adjustment strategies from LoRA+ into Serial LoRA, demonstrating its compatibility with existing LoRA improvements and highlighting the extensibility of our approach. These systematic evaluations across diverse vision tasks and model architectures validate the efficiency and versatility of Serial LoRA as a general parameter-efficient fine-tuning method.
\begin{figure*}[!t]
    \centering
    \setlength{\abovecaptionskip}{-0.2cm}
    \setlength{\belowcaptionskip}{-0.3cm}
    \includegraphics[width=\linewidth]{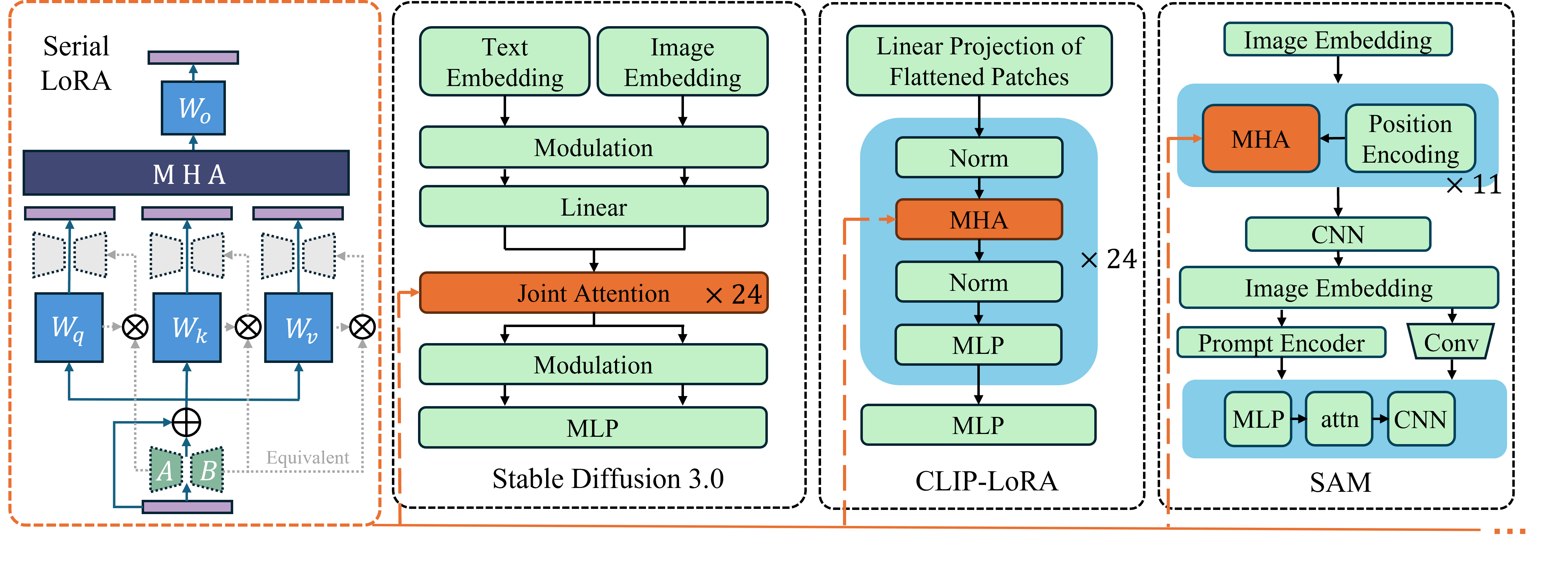}
    \caption{Instead of learning separate pairs of matrices, Serial LoRA learns a shared pair of low-rank matrices, significantly reducing the training parameter requirements. Its strong scalability allows it to be directly applied to various vision tasks, such as CLIP, Stable Diffusion 3.0 and SAM, enhancing efficiency across diverse applications. }
    \label{fig:lora}
\end{figure*}

In summary, our contributions are as follows: 
\begin{itemize}
\item  We propose Serial LoRA, an novel method that leverages a shared low-rank matrix serially composite with the attention mechanisms, which equivalently builds multiple LoRAs respectively adapted with different parameters.

\item Serial LoRA can be seamlessly integrated into the adaptation of various vision foundation models with the transformer structure, expanding the scope of LoRA for more efficient fine-tuning in resource-constrained scenarios.

\item  Extensive experiments on 24 datasets encompassing classification, segmentation and generation, consistently demonstrate the comparable performance of Serial LoRA with 1/4 parameters of LoRA in most cases. 

\end{itemize}

\section{Methods}

\subsection{preliminary}
Without loss of generality, for a pre-trained weight matrix $\mathbf{W}_0 \in \mathbb{R}^{d_1 \times d_2}$ where $d_1$ and $d_2$ are the dimensions of the parameter layer, LoRA models the difference between the pre-trained and fine-tuned weights as the product of two low-rank matrices:
\begin{align}
\widebar{\mathbf{W}} = \mathbf{W}_0 + \Delta \mathbf{W} = \mathbf{W}_0 + \mathbf{B} \mathbf{A}    
\end{align}
where $\mathbf{B} \in \mathbb{R}^{d_1 \times r}$, $\mathbf{A} \in \mathbb{R}^{r \times d_2}$, and $r \ll \min\{d_1,~d_2\}$.
During fine-tuning, $\mathbf{W}_0$ remains fixed while only $\Delta \mathbf{W}$ undergoes training. Matrix $\mathbf{A}$ is initialized with random Gaussian noise, ensuring diverse updates, while $\mathbf{B}$ is initialized to zero, setting $\Delta \mathbf{W} = 0$ at the begining, which avoids immediate interference with the pre-trained weights.
This low-rank adaptation approach quickly gained traction in large vision models, especially within the context of vision transformers, leading to the development of various extensions such as LoRA+\cite{hayou2024loraefficientlowrank}, and DoRA\cite{liu2024doraweightdecomposedlowrankadaptation}. These methods commonly adapt LoRA to the query ($q$), key ($k$), value ($v$), and output ($out$) projection matrices in the attention block.
Formally, for any projection matrix $\mathbf{W}_{\text{proj}} \in \{\mathbf{W}_q, \mathbf{W}_k, \mathbf{W}_v, \mathbf{W}_{\text{out}}\}$, the fine-tuned weight $\widebar{\mathbf{W}_{\text{proj}}}$ is defined by:
\begin{align} 
\widebar{\mathbf{W}_{\text{proj}}} &= \mathbf{W}_{\text{proj}} + \Delta \mathbf{W}_{\text{proj}} = \mathbf{W}_{\text{proj}} + \mathbf{B}_{\text{proj}} \mathbf{A}_{\text{proj}}
\end{align}
By concentrating on essential weight updates, LoRA effectively adapts large-scale vision transformers to new tasks.

\subsection{motivation}
Despite rapid advancements, current LoRA techniques still fall short of meeting real-world demands, especially where the cost of fine-tuning remains high due to substantial parameter requirements. 
Although several methods have further optimized its parameter redundancies, none of them has considered the inherent network structure, e.g. the vision transformer.
This raises us a question: rather than transferring large models with extensive parameters encoding prior knowledge to personalized domains, could we instead learn to adapt the commonality among parameters in vision transformer to further compress the parameter redundancy? To address this, we propose exploring a unified low-rank space within the multi-head attention (MHA) structure of Transformer architectures. By sharing adaptive parameters, we can further reduce the number of trainable parameters to be optimized, enabling more scalable adaption to various resource-constraint tasks and domains.

\begin{figure}[h!]
    \centering
    \setlength{\belowcaptionskip}{-0.3cm}
    \includegraphics[width=\linewidth]{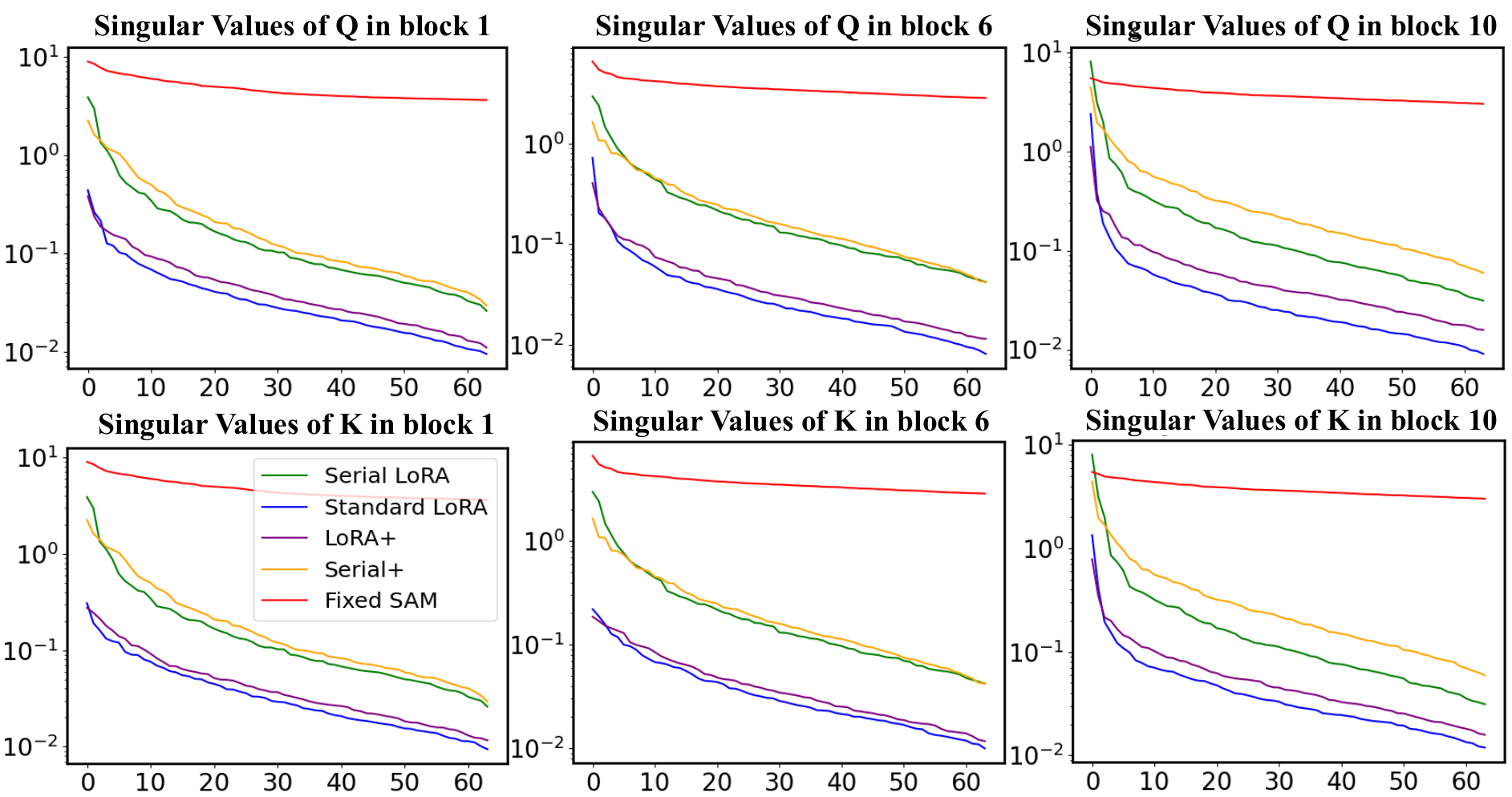}
    \caption{Singular value analysis varying different transformer blocks in SAM model. The comparison between SAM weights (red), Serial LoRA (green), LoRA (blue), LoRA+ (purple), and Serial LoRA+ (orange) demonstrates that Serial LoRA variants maintain intermediary singular value distributions with gradual decay patterns across different network depths.}
    \label{fig:svd}
\end{figure}

\subsection{Serial LoRA for Vision Transformer}
Given the aforementioned discussion, we propose \textbf{Serial LoRA}, a parameter-efficient fine-tuning approach that introduces a shared low-rank transformation in vision transformers. Serial LoRA learns a pair of low-rank matrices, \( \mathbf{A}_s \in \mathbb{R}^{r \times d_2} \) and \( \mathbf{B}_s \in \mathbb{R}^{d_1 \times r} \), to directly transform input features for adaptation within the model's pre-trained parameter space. We define the fine-tuning weight matrix as \( \Delta \mathbf{W}_s = \mathbf{B}_s \mathbf{A}_s \) and apply it on the input feature \( \mathbf{x} \) of MHA block to generate the adapted query, key, and value embeddings.
Take query embedding as example, we have:
\begin{align}
\tilde{\mathbf{q}} &= \mathbf{W}_q (\mathbf{I} +  \mathbf{B}_s \mathbf{A}_s) \mathbf{x}
\end{align}
where $\tilde{\mathbf{q}}$ is the adapted query, 
The same fine-tuning process applies to the key and value embeddings, with analogous expressions for $\tilde{\mathbf{k}},\tilde{\mathbf{v}}$.
These matrices encapsulate the model's pre-trained knowledge and serve as the foundation for our task-specific adaptations.
This design differs fundamentally from conventional methods in two aspects. First, while standard LoRA applies separate learnable matrices in parallel for each attention component, Serial LoRA learns a single, shared adaptation matrix that acts as a common adjustment. Second, this transformation is applied before the projection through pre-trained weights, enabling a more uniform adaptation across attention components.The shared nature of \( \mathbf{B}_s \mathbf{A}_s \) not only significantly reduces the number of trainable parameters but also aligns the adaptations within a unified parameter space. This serial propagation of a shared transformation achieves parameter efficiency and maintains model adaptability through a more compact and unified design. 

\noindent\textbf{Extension.} We can easily integrate Serial LoRA into diverse vision foundation models with the transformer structure. 
The detailed architecture modifications and integration strategy of Serial LoRA are illustrated in Fig. \ref{fig:lora}, which demonstrates how our approach can be seamlessly incorporated into various transformer-based models. In the experimental part, we will demonstrate that Serial LoRA learns a unified, parameter-efficient way for adapting vision transformers efficiently across diverse architectures and tasks. 

\subsection{Analysis}
\noindent\textbf{Equivalent Form.} Without loss of generality, we take the query embedding as example. With a pre-trained matrix \( \mathbf{W}_q \), both Serial LoRA and standard LoRA adapt to new parameter spaces by learning transformations of input features in the attention mechanism. Due to the different positions and strategies applied in the transformer block, standard LoRA and Serial LoRA exhibit distinct  formulations in their transformations, as shown in (\ref{eq:standard_lora}) and (\ref{eq:serial_lora}):
\begin{align}
\mathbf{q} &= (\mathbf{W}_q + \mathbf{B}_q \mathbf{A}_q) \mathbf{x} = \mathbf{W}_q \mathbf{x} + \mathbf{B}_q \mathbf{A}_q \mathbf{x}, \label{eq:standard_lora} \\
\tilde{\mathbf{q}} &= \mathbf{W}_q (\mathbf{I} +  \mathbf{B}_s \mathbf{A}_s) \mathbf{x} = \mathbf{W}_q \mathbf{x} + \mathbf{W}_q \mathbf{B}_s \mathbf{A}_s \mathbf{x}. \label{eq:serial_lora}
\end{align}
Standard LoRA directly transform the input by \( \mathbf{B}_q \mathbf{A}_q \), but Serial LoRA first forms a composite matrix by \( \mathbf{W}_q \mathbf{B}_s \mathbf{A}_s \). It uses the shared matrix to be composite with the weight matrix in attention mechanism, which individually builds different adaptation based on the basis $\textbf{W}_q$ ($\textbf{W}_k$ or $\textbf{W}_v$). From this perspective, our method actually adapt like LoRA by means of the specification of different basis $\textbf{W}$.

\noindent\textbf{Non-equivalent Dynamic.} Although there are some potential equivalent form between Serial LoRA and LoRA, we should point out that they perform very different. To be clear, in Fig. \ref{fig:svd}, we conducted singular value analysis on the Q, K, and V projection matrices across different transformer blocks learned by Serial LoRA and LoRA. As can be seen, Serial LoRA demonstrates two key properties in its singular value distribution: a consistent intermediate positioning between the original pre-trained model and standard LoRA, and a more gradual decay pattern compared to the steep deterioration in standard LoRA. The intermediate positioning suggests that Serial LoRA achieves an optimal balance between parameter efficiency and representation capacity. Specifically, the higher singular values compared to standard LoRA indicate better preservation of the model's original representation power, while still maintaining significant parameter reduction relative to the original model. The gradual decay pattern further implies that Serial LoRA retains a broader spectrum of feature dimensions, contributing to more robust feature representations. These characteristics persist across both different attention components (Q, K, and V matrices) and various transformer blocks from shallow to deep layers. This consistent behavior throughout the network hierarchy demonstrates that Serial LoRA's enhanced representation capacity is systematically maintained, likely contributing to its superior performance across diverse downstream tasks.
\section{Experiments}

To validate the effectiveness of our method, we conduct extensive experiments on 24 datasets across three major vision tasks: image generation using Stable Diffusion 3.0~\cite{SD3}, image classification with CLIP~\cite{pmlr-v139-radford21a}, and semantic segmentation with SAM~\cite{Kirillov_2023_ICCV}. Our method, Serial LoRA, is compared directly with LoRA~\cite{hu2022lora}. Additionally, we extend Serial LoRA into the LoRA+~\cite{hayou2024loraefficientlowrank} framework and perform further comparisons with LoRA+ to evaluate its compatibility and performance within advanced fine-tuning structures. Please refer to the supplementary material for more results.
\begin{table}[h]
\centering
\resizebox{0.5\textwidth}{!}{%
\begin{tabular}{ccc}
\toprule
\toprule

               & LoRA / LoRA+            & Serial LoRA / Serial LoRA+\\ \hline
CLIP~\cite{pmlr-v139-radford21a}           & 3.84M   & \textbf{0.96M}        \\ 
SAM~\cite{Kirillov_2023_ICCV} & 18.87M             & \textbf{4.72M}\\ 
SD3~\cite{SD3}            & 19.32M &  \textbf{4.83M }     \\
\bottomrule
\bottomrule
\end{tabular}
}
\caption{Comparison of Parameter Counts Between LoRA  and Serial LoRA. Our method achieves a 4-fold reduction in parameters compared to LoRA.}
\label{tab:size}
\end{table}

\subsection{Serial Lora on Diffusion}
\textbf{Experimental Setup}.
For our generative tasks, we evaluate eight stylistic datasets: Barbie, Cyberpunk, Art Nouveau, Impressionism. Barbie and Cyberpunk contain 315 and 439 images, respectively,\footnote{\url{https://github.com/sjtuplayer/SaRA}.} with prompts drawn from captions, while the remaining datasets each include 1,500 images using artwork titles as prompts.\footnote{\url{https://github.com/liaopeiyuan/artbench}.} According to ~\cite{Wu_2025_WACV}, we assess quality using Fréchet Inception Distance (FID) and CLIP-Score. We utilize Stable Diffusion 3.0 \cite{SD3} for image generation, comparing LoRA, Serial LoRA, LoRA+, and Serial LoRA+ to assess compatibility and parameter efficiency.  Training is set to 10 epochs per dataset with an initial learning rate of $1.5 \times 10^{-5}$. For LoRA+ and Serial LoRA+, we maintain a learning rate ratio of 1:20 between matrices $\mathbf{A}$ and $\mathbf{B}$. All method is configured with a default rank of 64. Evaluation involves 28 inference steps at a $1024 \times 1024$ resolution, with classifier-free guidance weight $\omega=7.0$ to ensure consistency across methods. All experiments run on a single Nvidia RTX 3090 GPU in bfloat16 precision.
\begin{table}[!h]
\resizebox{0.5\textwidth}{!}{%
\setlength{\tabcolsep}{2pt}
\begin{tabular}{cccccccccccccccccc}

\toprule
\toprule
                           &              & \multicolumn{2}{c}{\multirow{2}{*}{Barbie}}              & \multicolumn{2}{c}{\multirow{2}{*}{Cyberpunk}}         & \multicolumn{2}{c}{\multirow{2}{*}{\makecell{Art \\ Nouveau}}}                             & \multicolumn{2}{c}{\multirow{2}{*}{\makecell{Impressionism}}}                   \\ \\ \hline
                           &              & \multirow{2}{*}{FID$\downarrow$}  & \multirow{2}{*}{\makecell{CLIP\\Score}$\uparrow$} & \multirow{2}{*}{FID$\downarrow$} & \multirow{2}{*}{\makecell{CLIP\\Score}$\uparrow$} &\multirow{2}{*}{FID$\downarrow$} & \multirow{2}{*}{\makecell{CLIP\\Score}$\uparrow$} & \multirow{2}{*}{FID$\downarrow$} & \multirow{2}{*}{\makecell{CLIP\\Score}$\uparrow$}    \\ \\ \hline
\multirow{4}{*}{\makecell{20\% \\Shot}} & LoRA         & \textbf{189.62}  & 0.8223               &\textbf{ 172.43 }          & \textbf{0.8146}      & 153.31           & \textbf{0.6686}            & 153.98           & 0.6963                              \\
                           & Serial LoRA  & 193.66           & \textbf{0.8302}      & 173.54  & 0.8135               & \textbf{152.32}  & 0.6683                              & \textbf{153.30}  & \textbf{0.7003}            \\ \cdashline{3-18}[2pt/5pt]
                           & LoRA+        & 187.21  & 0.8122               & 173.68           & 0.8127               & \textbf{152.83}  & 0.6609                     & 153.81  & 0.6983            \\
                           & Serial LoRA+ & \textbf{182.31 }          & \textbf{0.8222}      & \textbf{172.21}  & \textbf{0.8178}      & 153.04           & \textbf{0.6693}            & \textbf{153.79}           & \textbf{0.6998}            \\ \hline
\multirow{4}{*}{\makecell{50\% \\Shot}} & LoRA         & 142.16           & 0.8256      & 115.88           & \textbf{0.8235}      & \textbf{94.62}   & 0.6795                     & 102.92           & 0.7051          \\
                           & Serial LoRA  & \textbf{141.93}  & \textbf{0.8264}      & \textbf{115.62}  & 0.8227      & 94.87            & \textbf{0.6841}            & \textbf{102.65}  & \textbf{0.7052}                              \\ \cdashline{3-18}[2pt/5pt]
                           & LoRA+        & 142.79  & 0.8243               & \textbf{116.25}  & 0.8231               & 94.63            & 0.6815            & \textbf{102.10}           & 0.7020                     \\
                           & Serial LoRA+ & \textbf{142.14}           & \textbf{0.8251}      & 117.41           & \textbf{0.8232}      & \textbf{94.57}   & \textbf{0.6884}            & 102.33  & \textbf{0.7038}                    \\ \hline
\multirow{4}{*}{\makecell{Full\\ Shot}} & LoRA         & \textbf{116.35}  & 0.8298      & \textbf{99.39}   & 0.8282               & \textbf{67.25}            & 0.6786                     & \textbf{73.50}   & 0.7010                             \\
                           & Serial LoRA  & 117.93           & \textbf{0.8302}      & 99.88            & \textbf{0.8285}      & 67.29   & \textbf{0.6791}            & 74.08            & \textbf{0.7030}            \\ \cdashline{3-18}[2pt/5pt]
                           & LoRA+        & 115.51           & 0.8294      & 99.85   & 0.8279               & 68.04   & 0.6792                              & \textbf{72.81}   & 0.7020                     \\
                           & Serial LoRA+ & \textbf{114.42}  &\textbf{ 0.8299}               & \textbf{99.27}           & \textbf{0.8290}      & \textbf{66.87}            & \textbf{0.6794}            & 74.10            & \textbf{0.7031}                     \\ 
\bottomrule
\bottomrule
\end{tabular}

}
\caption{Quantitative Comparison of FID and CLIP-Score Between Serial LoRA and LoRA on the Stable Diffusion 3.0, with Model Parameters Reduced from 19.32M to \textbf{4.83M}.}
\label{tab:sd3}
\end{table}

\textbf{Quantitative Results.} Tab. \ref{tab:sd3} presents a quantitative comparison of FID and CLIP-Score between our proposed Serial LoRA and Serial LoRA+ methods against standard LoRA and LoRA+ on the Stable Diffusion 3.0 model. The results demonstrate that Serial LoRA and Serial LoRA+ achieve comparable performance to LoRA and LoRA+ while using only 1/4 of the parameters, significantly reducing the model's memory footprint from 19.32M to 4.83M parameters. Specifically, when comparing Serial LoRA to LoRA, we observe that Serial LoRA maintains similar or even improved FID and CLIP-Score across most styles, such as Barbie and Art Nouveau, indicating strong generative quality with lower parameter costs. Similarly, Serial LoRA+ shows robust performance relative to LoRA+, achieving comparable or superior scores in various styles like Cyberpunk and Impressionism. These findings confirm that our method integrates seamlessly with existing LoRA structures, significantly reducing parameter overhead while retaining high-quality outputs across diverse styles. 

\begin{figure}[h]
    \centering
    \includegraphics[width=0.5\textwidth]{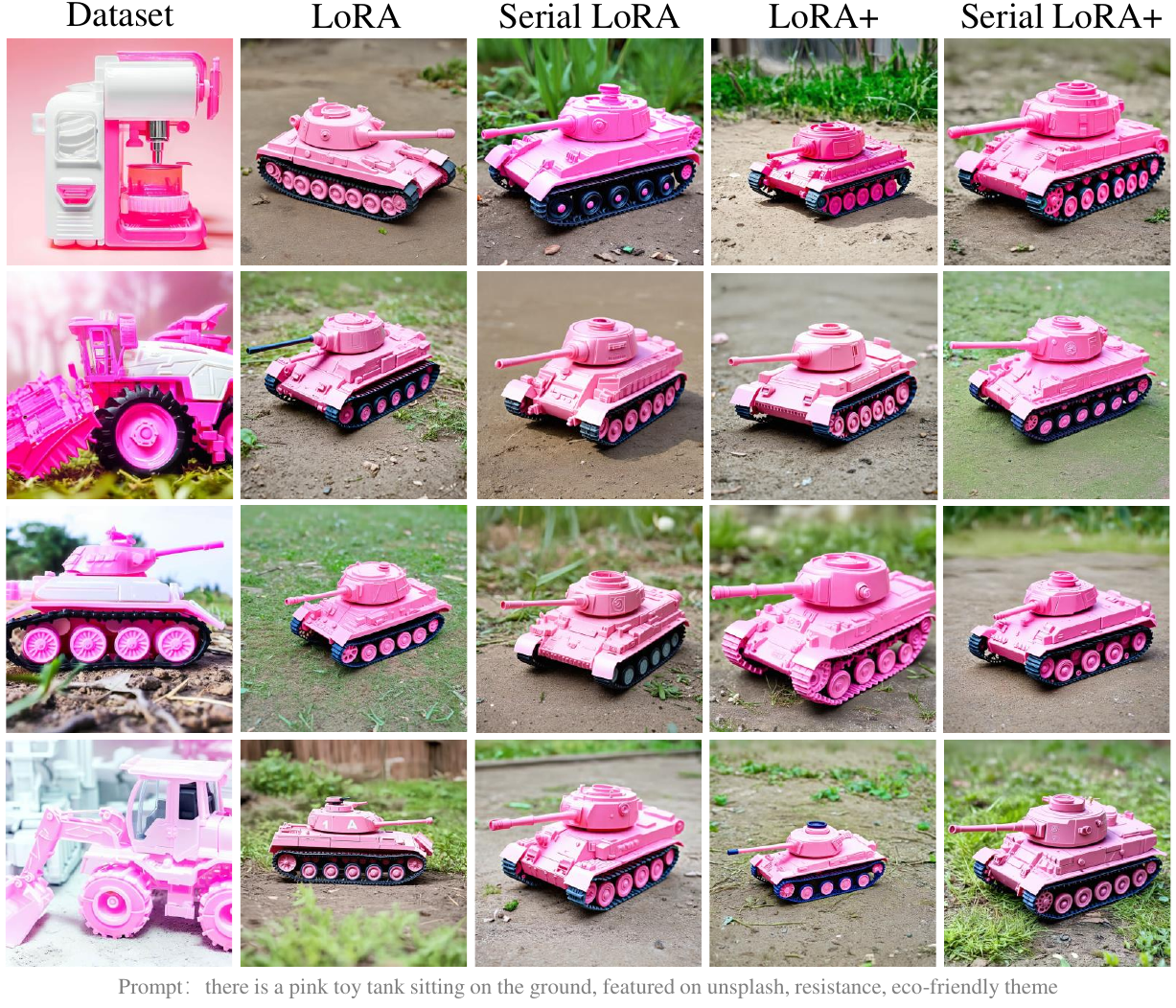}
    \caption{Qualitative Results about the comparison of Serial LoRA and LoRA. }
    \label{fig:serial_lora_diffusion}
\end{figure}

\textbf{Qualitative Results.} We use a consistent prompt as input to generate multiple images with specific styles. Part of the qualitative comparison results of LoRA, Serial LoRA, LoRA+, and Serial LoRA+ are shown in Fig. \ref{fig:serial_lora_diffusion}.  It can be seen that our model can learn the style accurately while generating images that align well with the given text prompts across different datasets. These findings further validate that Serial LoRA integrates effectively with existing Transformer-based diffusion models, enabling efficient fine-tuning and the creation of high-quality, distinctively styled megapixel images.

\subsection{Serial Lora on CLIP}

\textbf{Experimental Setup}
For the image classification task, we evaluate the performance of Serial LoRA on the CLIP model (ViT-B/16) \cite{pmlr-v139-radford21a} across eight benchmark datasets: Caltech101 (101 classes) \cite{Caltech-101}, Food 101 (101 classes) \cite{Food101}, EuroSAT (10 classes) \cite{eurosat} and Oxford Pets (37 classes) \cite{Oxford-pet}. Average accuracy is used as the primary metric to assess model performance. Each model is trained for 50 epochs, with a fixed learning rate of $2$e-$4$, and a dropout rate of 0.25. All method is configured with a default rank of 32.For LoRA+ and Serial LoRA+, we maintain a learning rate ratio of 1:20 between matrices $\mathbf{A}$ and $\mathbf{B}$. All experiments are conducted on a single Nvidia RTX 3090 GPU in bfloat16.

\begin{table}[h]
\resizebox{0.5\textwidth}{!}{%
\setlength{\tabcolsep}{2pt}
\begin{tabular}{cccccc}
\toprule
\toprule
                         &              & Caltech101     & Food 101       & EuroSAT        & OxFord Pets    \\ \hline
\multirow{4}{*}{1 Shot}  & LoRA         & 93.22   & 83.64   & 69.90    & 87.87    \\
                         & Serial LoRA  & \textbf{93.69}& \textbf{85.77}& \textbf{70.80}    & \textbf{90.52}   
\\ \cdashline{3-6}[2pt/5pt]
                         & LoRA+        & 93.31      & 84.08  & 59.45   & 88.23  
\\
                         & Serial LoRA+ & \textbf{94.44}     & \textbf{85.88}     & \textbf{67.36}& \textbf{89.59}\\ \hline
\multirow{4}{*}{4 Shot}  & LoRA         & 94.51      & 84.08    & 86.42     & 90.13     
\\
                         & Serial LoRA  & \textbf{95.54}     & \textbf{86.04}    & \textbf{86.98}& \textbf{91.92}    
\\ \cdashline{3-6}[2pt/5pt]
                         & LoRA+        & 92.60       & 79.44  & 76.01 & 85.78     
\\
                         & Serial LoRA+ & \textbf{94.48}& \textbf{82.79}& \textbf{88.25}    & \textbf{89.51}\\ \hline
\multirow{4}{*}{16 Shot} & LoRA         & 96.25      & 84.69      & \textbf{92.75}  & 92.45     
\\
                         & Serial LoRA  & \textbf{96.27}     & \textbf{86.98}    & 91.90      & \textbf{92.95}    
\\ \cdashline{3-6}[2pt/5pt]
                         & LoRA+        & 95.02      & 77.98     & 88.11     & 87.78  
\\
                         & Serial LoRA+ & \textbf{95.94}& \textbf{81.92}& \textbf{92.12}& \textbf{89.07}\\
\bottomrule
\bottomrule
\end{tabular}
}
\caption{Comparison of Classification Accuracy(\%) between Serial LoRA and LoRA on the CLIP Method, with Model Parameters Reduced from 3.84M to \textbf{0.96M}.}
\label{tab:clip}
\end{table}
\textbf{Analysis}. Tab. \ref{tab:clip} presents a comparison of classification accuracy between Serial LoRA, Serial LoRA+, and standard LoRA and LoRA+ methods across eight benchmark datasets using the CLIP model. The results demonstrate that our proposed Serial LoRA and Serial LoRA+ methods achieve comparable or even superior accuracy to LoRA and LoRA+ while significantly reducing model parameters from 3.84M to 0.96M. Specifically, in few-shot learning scenarios, particularly in the 1 Shot setting, Serial LoRA consistently exhibits marginally improved accuracy across almost all datasets.  Notably, Serial LoRA+ outperforms LoRA+ in accuracy across all datasets and shot settings, confirming its efficacy in achieving parameter efficiency without compromising performance. This improvement can be attributed to the design of Serial LoRA, which leverages a shared low-rank matrix in a serial configuration. This structure captures parameter commonalities across the attention heads more effectively, reducing redundancy while maintaining expressive power.

\subsection{Serial LoRA on SAM}
\textbf{Experimental Setup}

For our segmentation tasks, we construct eight few-shot segmentation datasets across diverse categories: one flood segmentation task\footnote{\url{https://datasetninja.com/floodnet}.}, human action segmentation task\footnote{\url{http://vision.stanford.edu/Datasets/40actions.html}.} and three food category segmentation tasks\footnote{\url{https://github.com/LARC-CMU-SMU/FoodSeg103-Benchmark-v1}.}.  Segmentation performance is measured by the Dice Score\cite{diceloss}, where a lower score indicates greater accuracy. For implementation, we use the SAM-ViT-B backbone, fine-tuning only the Vision Transformer (ViT) in the image encoder, while keeping the prompt encoder and mask decoder frozen. Parameter efficiency comparisons are conducted with a rank of 64 for all methods, using an initial learning rate of $1\times10^{-4}$ and a 1:20 ratio between matrices $\mathbf{A}$ and $\mathbf{B}$ for LoRA+ variants. Models are trained for 5 epochs across all datasets on an Nvidia RTX 3090 GPU in bfloat16 precision.

\textbf{Analysis}. Tab. \ref{tab:sam} indicates that Serial LoRA and Serial LoRA+ outperform LoRA and LoRA+ in nearly all cases, despite using only \textbf{1/4} of the original parameter count. For instance, Serial LoRA shows a slight improvement over LoRA across various shot settings on datasets such as Flood and Strawberry. Additionally, the DICE score indicates that Serial LoRA yields more accurate segmentation masks. Specifically, Serial LoRA+ achieves state-of-the-art results in few-shot learning involving comparatively fewer samples, demonstrating both the compatibility of our approach and the superior performance of Serial LoRA under more limited data conditions.  These findings demonstrate that Serial LoRA can be smoothly incorporated into existing LoRA architectures, enabling more accurate segmentation even with significantly reduced trainable parameters.

\begin{table}[!h]
\resizebox{0.5\textwidth}{!}{%
\begin{tabular}{ccccccc}
\toprule
\toprule
                           &              & Flood           & Human           & Icecream        & Pie             & Strawberry      \\ \hline
\multirow{4}{*}{20\% Shot} & LoRA         & 0.2983          & 0.0943          & \textbf{0.2079} & \textbf{0.2483} & 0.2736          \\ 
                           & Serial LoRA  & \textbf{0.2721} & \textbf{0.0790}  & 0.2214          & 0.3376          & \textbf{0.1675} \\ \cdashline{3-7}[2pt/5pt]& LoRA+        & \textbf{0.2469} & 0.1575          & 0.2464          & 0.3560          & 0.4512          \\
                           & Serial LoRA+ & 0.2509          & \textbf{0.1574} & \textbf{0.2413} & \textbf{0.1822} & \textbf{0.4490} \\ \hline
\multirow{4}{*}{50\% Shot} & LoRA         & 0.2487          & 0.0806          & 0.2227          & 0.2910          & 0.1609          \\
                           & Serial LoRA  & \textbf{0.2355} & \textbf{0.0697} & \textbf{0.1887} & \textbf{0.2803} & \textbf{0.1344} \\ \cdashline{3-7}[2pt/5pt]& LoRA+        & 0.2659          & 0.0971          & 0.2747          & 0.3323          & 0.3013          \\
                           & Serial LoRA+ & \textbf{0.2478} & \textbf{0.0878} & \textbf{0.2370} & \textbf{0.1809} & \textbf{0.2452} \\ \hline
\multirow{4}{*}{Full Shot} & LoRA         & 0.2822          & 0.0618          & 0.1673          & 0.2585          & 0.1140          \\
                           & Serial LoRA  & \textbf{0.2333} & \textbf{0.0439} & \textbf{0.1642} & \textbf{0.1856} & \textbf{0.1116} \\ \cdashline{3-7}[2pt/5pt]& LoRA+        & 0.2419          & \textbf{0.0692} & 0.1778          & 0.2898          & 0.1453          \\
                           & Serial LoRA+ & \textbf{0.2411} & 0.0727          & \textbf{0.1717} & \textbf{0.1551} & \textbf{0.1383} \\ 
\bottomrule
\bottomrule
\end{tabular}
}
\caption{Dice$\downarrow$ score comparison between Serial LoRA (\textbf{4.72MB}) and LoRA (18.87MB) for SAM-based few-shot segmentation tasks.}
\label{tab:sam}
\end{table}

As shown in Tab. \ref{tab:size} For the experiments conducted on CLIP, SAM, and SD3 models, our shared parameter design in Serial LoRA ensures a structural parameter reduction of 1/4 compared to the standard LoRA approach.

\subsection{Further Study}
\begin{figure}[ht]
    \vspace{-4mm}
    \centering
    \includegraphics[width=\linewidth]{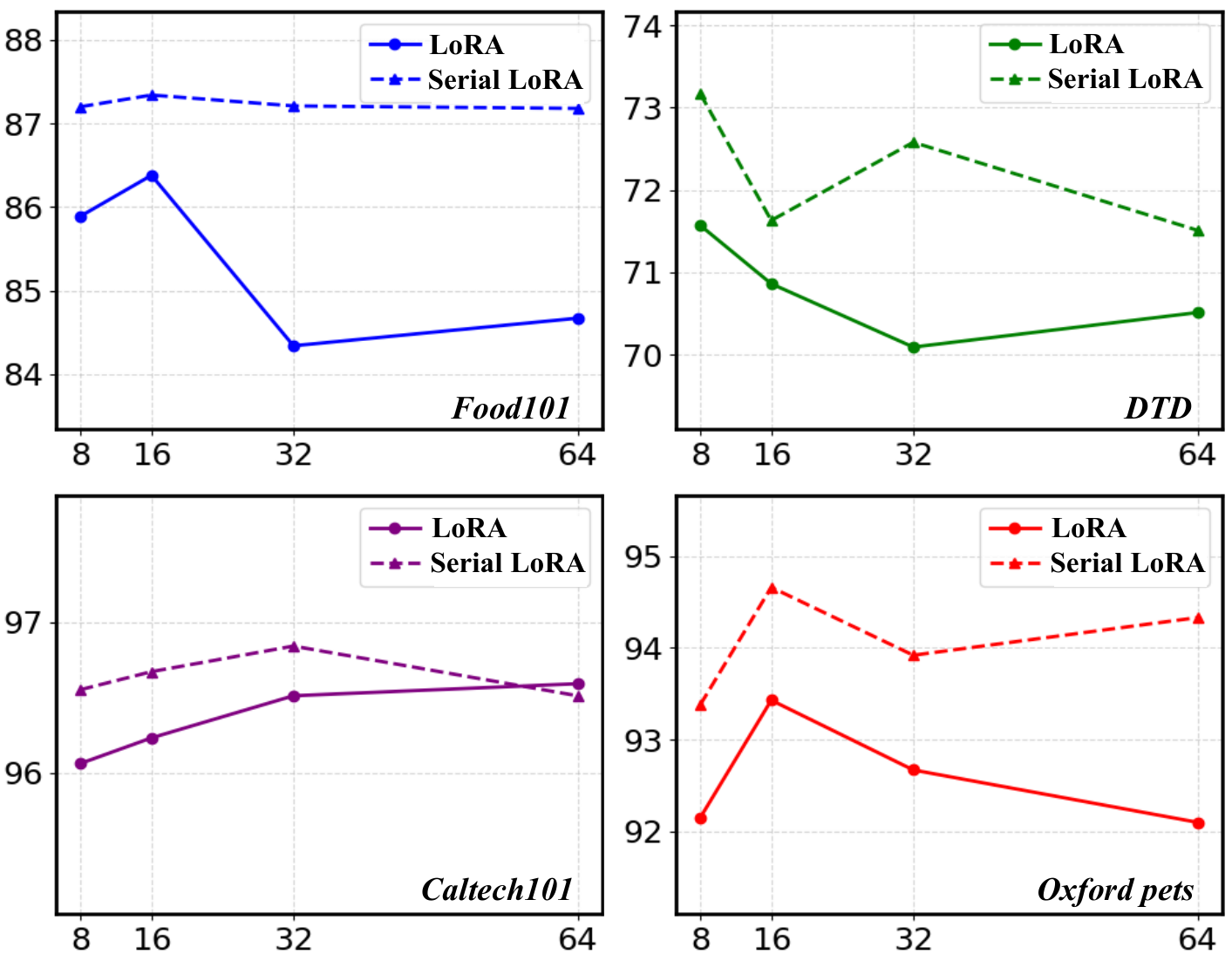}
    \vspace{-3mm}
    \caption{Comparison of LoRA and Serial LoRA performance in the CLIP model on image classification tasks from rank 8 to 64, measured in accuracy (\%). The datasets used include Food101, DTD, Caltech101, and Oxford Pets.}
    \label{fig:cliprankabla}
\end{figure}

We validate the adaptability of the Serial LoRA method under different rank settings and compare its performance with standard LoRA. For image classification with CLIP (Fig. \ref{fig:cliprankabla}), we evaluate Serial LoRA by varying the rank from 8 to 64 across diverse datasets including Food101, DTD, Caltech101, and Oxford Pets.
With identical rank settings, Serial LoRA reduces the parameter count to 1/4 of standard LoRA through shared transformation design.

\vspace{-4mm}
\section{Conclusion}

In this paper, we propose \textbf{Serial LoRA}, a novel parameter-efficient fine-tuning method, tailored for various computer vision tasks. By identifying the shared parameter space of the multi-head attention(MHA) block within pretrained Transformers and applying a consistent low-rank adaptation approach, we significantly reduce the number of fine-tuning parameters. In experiments, we applied Serial LoRA to image generation, image classification, and semantic segmentation tasks. With fine-tuning parameters reduced to only \textbf{1/4} of the original, Serial LoRA demonstrated the effectiveness of our method while minimally impacting the model's performance. Furthermore, our approach is compatible with existing PEFT methods, underscoring the broad applicability of Serial LoRA.

\bibliographystyle{IEEEbib}
\bibliography{icme2025_template_anonymized.bbl}

\begin{thebibliography}{10}

\bibitem{Kirillov_2023_ICCV}
Alexander Kirillov, Eric Mintun, Nikhila Ravi, Hanzi Mao, Chloe Rolland, Laura Gustafson, Tete Xiao, Spencer Whitehead, Alexander~C. Berg, Wan-Yen Lo, Piotr Dollar, and Ross Girshick,
\newblock ``Segment anything,''
\newblock in {\em Proceedings of the IEEE/CVF ICCV}, 2023, pp. 4015--4026.

\bibitem{pmlr-v139-radford21a}
Alec Radford, Jong~Wook Kim, Chris Hallacy, Aditya Ramesh, Gabriel Goh, Sandhini Agarwal, Girish Sastry, Amanda Askell, Pamela Mishkin, Jack Clark, Gretchen Krueger, and Ilya Sutskever,
\newblock ``Learning transferable visual models from natural language supervision,''
\newblock in {\em Proceedings of the 38th ICML}, 2021, vol. 139, pp. 8748--8763.

\bibitem{SD3}
Patrick Esser, Sumith Kulal, Andreas Blattmann, Rahim Entezari, Jonas M{\"u}ller, Harry Saini, Yam Levi, Dominik Lorenz, Axel Sauer, Frederic Boesel, Dustin Podell, Tim Dockhorn, Zion English, and Robin Rombach,
\newblock ``Scaling rectified flow transformers for high-resolution image synthesis,''
\newblock in {\em Proceedings of the 41th ICML}, 2024.

\bibitem{zhu2024minigpt}
Deyao Zhu, Jun Chen, Xiaoqian Shen, Xiang Li, and Mohamed Elhoseiny,
\newblock ``Mini{GPT}-4: Enhancing vision-language understanding with advanced large language models,''
\newblock in {\em The Twelfth International Conference on Learning Representations}, 2024.

\bibitem{dai2023instructblip}
Wenliang Dai, Junnan Li, Dongxu Li, Anthony Tiong, Junqi Zhao, Weisheng Wang, Boyang Li, Pascale Fung, and Steven Hoi,
\newblock ``Instruct{BLIP}: Towards general-purpose vision-language models with instruction tuning,''
\newblock in {\em the 37th NeurIPS}, 2023.

\bibitem{Benchmarking}
Shuo Chen, Jindong Gu, Zhen Han, Yunpu Ma, Philip Torr, and Volker Tresp,
\newblock ``Benchmarking robustness of adaptation methods on pre-trained vision-language models,''
\newblock in {\em the 37th NeurIPS Datasets and Benchmarks Track}, 2023.

\bibitem{huetal2023llm}
Zhiqiang Hu, Lei Wang, Yihuai Lan, Wanyu Xu, Ee-Peng Lim, Lidong Bing, Xing Xu, Soujanya Poria, and Roy Lee,
\newblock ``{LLM}-adapters: An adapter family for parameter-efficient fine-tuning of large language models,''
\newblock in {\em Proceedings of the 2023 Conference on Empirical Methods in Natural Language Processing}, 2023, pp. 5254--5276.

\bibitem{adapterfusion}
Jonas Pfeiffer, Aishwarya Kamath, Andreas R{\"u}ckl{\'e}, Kyunghyun Cho, and Iryna Gurevych,
\newblock ``Adapterfusion: Non-destructive task composition for transfer learning,''
\newblock in {\em Proceedings of the 16th Conference of the European Chapter of the Association for Computational Linguistics: Main Volume}, 2021, pp. 487--503.

\bibitem{AdapterSoup}
Alexandra Chronopoulou, Matthew~E. Peters, Alexander Fraser, and Jesse Dodge,
\newblock ``Adaptersoup: Weight averaging to improve generalization of pretrained language models,''
\newblock in {\em EACL (Findings)}, 2023, pp. 2009--2018.

\bibitem{hu2022lora}
Edward~J Hu, yelong shen, Phillip Wallis, Zeyuan Allen-Zhu, Yuanzhi Li, Shean Wang, Lu~Wang, and Weizhu Chen,
\newblock ``Lo{RA}: Low-rank adaptation of large language models,''
\newblock in {\em International Conference on Learning Representations}, 2022.

\bibitem{li2024loftq}
Yixiao Li, Yifan Yu, Chen Liang, Nikos Karampatziakis, Pengcheng He, Weizhu Chen, and Tuo Zhao,
\newblock ``Loftq: Lo{RA}-fine-tuning-aware quantization for large language models,''
\newblock in {\em The Twelfth International Conference on Learning Representations}, 2024.

\bibitem{hayou2024loraefficientlowrank}
Soufiane Hayou, Nikhil Ghosh, and Bin Yu,
\newblock ``Lo{RA}+: Efficient low rank adaptation of large models,''
\newblock in {\em Forty-first International Conference on Machine Learning}, 2024.

\bibitem{Peng_2024_CVPR}
Zelin Peng, Zhengqin Xu, Zhilin Zeng, Lingxi Xie, Qi~Tian, and Wei Shen,
\newblock ``Parameter efficient fine-tuning via cross block orchestration for segment anything model,''
\newblock in {\em Proceedings of the IEEE/CVF Conference on CVPR}, June 2024, pp. 3743--3752.

\bibitem{li2024adapting}
Kevin Li and Pranav Rajpurkar,
\newblock ``Adapting segment anything models to medical imaging via fine-tuning without domain pretraining,''
\newblock in {\em AAAI 2024 Spring Symposium on Clinical Foundation Models}, 2024.

\bibitem{LCMLORA}
Simian Luo, Yiqin Tan, Suraj Patil, Daniel Gu, Patrick von Platen, Apolin{\'a}rio Passos, Longbo Huang, Jian Li, and Hang Zhao,
\newblock ``Lcm-lora: A universal stable-diffusion acceleration module,''
\newblock {\em arXiv preprint arXiv:2311.05556}, 2023.

\bibitem{Hartley_2024_CVPR}
Zane~K.J. Hartley, Rob~J. Lind, Michael~P. Pound, and Andrew~P. French,
\newblock ``Domain targeted synthetic plant style transfer using stable diffusion lora and controlnet,''
\newblock in {\em Proceedings of the IEEE/CVF Conference on CVPR Workshops}, June 2024, pp. 5375--5383.

\bibitem{smith2024continual}
James~Seale Smith, Yen-Chang Hsu, Lingyu Zhang, Ting Hua, Zsolt Kira, Yilin Shen, and Hongxia Jin,
\newblock ``Continual diffusion: Continual customization of text-to-image diffusion with c-lo{RA},''
\newblock {\em Transactions on Machine Learning Research}, 2024.

\bibitem{liu2024doraweightdecomposedlowrankadaptation}
Shih yang Liu, Chien-Yi Wang, Hongxu Yin, Pavlo Molchanov, Yu-Chiang~Frank Wang, Kwang-Ting Cheng, and Min-Hung Chen,
\newblock ``Do{RA}: Weight-decomposed low-rank adaptation,''
\newblock in {\em Forty-first International Conference on Machine Learning}, 2024.

\bibitem{Wu_2025_WACV}
Haoning Wu, Shaocheng Shen, Qiang Hu, Xiaoyun Zhang, Ya~Zhang, and Yanfeng Wang,
\newblock ``Megafusion: Extend diffusion models towards higher-resolution image generation without further tuning,''
\newblock in {\em Proceedings of the Winter Conference on Applications of Computer Vision (WACV)}, February 2025, pp. 3944--3954.

\bibitem{Caltech-101}
Li~Fei-Fei, Rob Fergus, and Pietro Perona,
\newblock ``Learning generative visual models from few training examples: An incremental bayesian approach tested on 101 object categories,''
\newblock {\em Computer vision and Image understanding}, vol. 106, no. 1, pp. 59--70, 2007.

\bibitem{Food101}
Lukas Bossard, Matthieu Guillaumin, and Luc Van~Gool,
\newblock ``Food-101 -- mining discriminative components with random forests,''
\newblock in {\em European Conference on Computer Vision}, 2014.

\bibitem{eurosat}
Patrick Helber, Benjamin Bischke, Andreas Dengel, and Damian Borth,
\newblock ``Eurosat: A novel dataset and deep learning benchmark for land use and land cover classification,''
\newblock {\em IEEE Journal of Selected Topics in Applied Earth Observations and Remote Sensing}, 2019.

\bibitem{Oxford-pet}
Omkar~M. Parkhi, Andrea Vedaldi, Andrew Zisserman, and C.~V. Jawahar,
\newblock ``Cats and dogs,''
\newblock {\em 2012 IEEE Conference on Computer Vision and Pattern Recognition}, pp. 3498--3505, 2012.

\bibitem{diceloss}
Fausto Milletari, Nassir Navab, and Seyed-Ahmad Ahmadi,
\newblock ``V-net: Fully convolutional neural networks for volumetric medical image segmentation,''
\newblock in {\em 2016 Fourth International Conference on 3D Vision (3DV)}, 2016, pp. 565--571.

\end{thebibliography}

\end{document}